\documentclass{article}
\usepackage{amssymb}
\usepackage{tipa}
\usepackage{spconf,amsmath,graphicx}
\usepackage{transparent}
\usepackage{import}
\usepackage{colortbl}
\usepackage{makecell}
\definecolor{Gray}{gray}{0.85}
\usepackage[table,xcdraw]{xcolor}

\usepackage{tabularx}

\title{Robust end-to-end deep audiovisual speech recognition}
\name{Ramon Sanabria$^{1}$\thanks{The authors would like to thank Susanne Burger for sharing her insights into the role of phonetics and phonology in an audio-visual setting.} \qquad Florian Metze$^{2}$ \qquad Fernando De La Torre$^{1}$}

\address{
  $^1$ Robotics Institute and $^2$ Language Technologies Institute\\
  Carnegie Mellon University\\
  Pittsburgh, PA; U.S.A.\\
  \texttt{\{ramons|fmetze|ftorre\}@andrew.cmu.edu}}

\begin{document}
\ninept
\maketitle
\begin{abstract}

Speech is one of the most effective ways of communication among humans. Even though audio is the most common way of transmitting speech, very important information can be found in other modalities, such as vision. Vision is particularly useful when the acoustic signal is corrupted. Multi-modal speech recognition however has not yet found wide-spread use, mostly because the temporal alignment and fusion of the different information sources is challenging.

This paper presents an end-to-end audiovisual speech recognizer (AVSR), based on recurrent neural networks (RNN) with a connectionist temporal classification (CTC)~\cite{graves2006connectionist} loss function. CTC creates sparse ``peaky'' output activations, and we analyze the differences in the alignments of output targets (phonemes or visemes) between audio-only, video-only, and audio-visual feature representations. We present the first such experiments on the large vocabulary IBM ViaVoice database, which outperform previously published approaches on phone accuracy in clean and noisy conditions.


\end{abstract}


\section{Introduction}
\label{sec:intro}



Although researchers have been trying to improve the performance of automatic speech recognition systems under noisy conditions for decades, the problem is far from being solved. Some solutions are focused on removing the noise from the signal or improving the feature representation of the audio channel. However, the amount of signal masked by additive noise presents a natural limitation to those solutions. For that reason, some researchers use an alternative non-related with audio modality, for example the visual channel. Vision is usually not affected by the acoustic environment, and thus immune to corruption by noise. It has been demonstrated by~\cite{mcgurk1976hearing} that humans also tend to put their attention to other information channels (i.e., vision) to ease the understanding of the speaker when the acoustic channel is corrupted. The addition of a new modality however into a temporal sequence classification problem creates several challenges. In addition to potentially having to estimate stream weights, the temporal alignment of both information sources is usually not constant. This is due to the nature of the speech production process, as well as the complexity of the technical solutions for transmission, storage, and encoding of audio-visual data. In this paper, we present a novel approach to audio-visual speech recognition, which will allow us to investigate these effects in new and interesting ways.

\begin{figure}
  \centering
  \def\svgwidth{\columnwidth}

  \graphicspath{{./figs/}}
  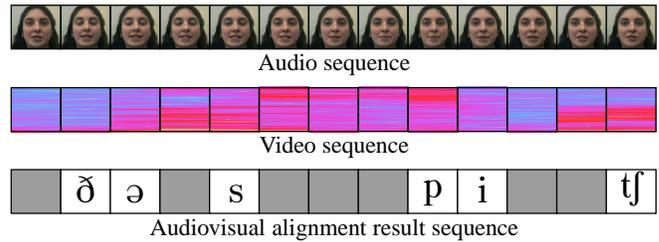
  \label{fig:alignment}
  \caption{Typical audio, video, and phonetic alignment sequence for the utterance ``the speech'' (phonetically transcribed into: \textipa{[D@ spitS]} using IPA notation), using CTC. 
  One frame corresponds to 33ms, the video frame rate.}
\end{figure}

Specifically, it is shown by different linguistic studies such as~\cite{kent1977coarticulation} that the mouth shape towards an articulatory target modify the following phone. This effect is accentuated when the speaking rate of the speaker is high~\cite{taylor2014effect}. In those cases, the visual modality will provide enough information to the system to determine how the phoneme should sound. However, a problem of synchronization between visemes (groups of similar movements of visual articulators) and phonemes is present. Some practical studies such as~\cite{bell1982temporal} state that the coarticulation is \textsl{speaker- and phoneme-dependent}. This adds a certain level of difficulty to the synchronization task between the audio and video modality, and makes frame-level fusion difficult. Consequently, somewhat a-synchronous approaches such as Hidden Markov models (HMMs) coupled at the state level have been attempted, but have also not met with dramatic success.

We present an AVSR solution that does not require an HMM, but rather uses several layers of bi-directional long short-term memory (LSTM)~\cite{hochreiter1997long} units as building blocks, followed by a CTC loss function for the output layer. The CTC loss is defined directly over the symbol sequence, and effectively marginalizes over all permitted alignments between frames and states, adding a ``blank'' state between label states. The resulting alignment typically contains mostly blanks, and is thus ``sparse'', see Fig.~\ref{fig:alignment}. In a multi-stream setting, where independent models are being trained and tested for the audio and video modality, frame-synchronous approaches such as tightly coupled HMMs together with score fusion (late integration) are therefore unlikely to work, because the ``peaks'' for the same unit will appear at a different point in time in each stream. Early integration (feature fusion) however should work just fine, and will thus be investigated in this paper.

As an end-to-end approach, CTC directly optimizes for the sequence of output labels without requiring any initial labeling (manual or ported from another system). However it is not clear if the \textsl{temporal locations} of the peaks have any meaningful interpretation. The audio-visual setting provides an interesting opportunity to investigate this issue: intuitively, the locations of the peaks should correspond to ``discriminative'' input features, and should thus mark ``informative'' time points. In an audio-visual setting, these may correspond to the different times at which phonemes and their corresponding visemes (we follow~\cite{cappelletta2012phoneme} for the mapping) are observed, without requiring any manual labeling or input.


This paper thus makes two main contributions: first, we demonstrate that CTC-based acoustic models can achieve state-of-the-art performance in audio-visual speech recognition tasks, as our system achieve comparable results with a traditional pre-Deep Learning baseline (~\cite{potamianos2001large} report 11\% Word Error Rate on the ViaVoice database) and outperforms recent cross-entropy trained DNN baseline ~\cite{mroueh2015deep} in terms of phoneme error rate. We report results for clean and noisy training and testing conditions. Our second contribution lies in an analysis of the peak structure of the CTC output labeling for a multi-modal input, which we can compare to human intuition about the nature and relationship between the feature generation process.


\section{Technical Background}
\label{sec:format}

\subsection{Architecture of AVSR systems}
\label{secsec:architecture}

Traditionally, HMMs and Gaussian Mixture Models (GMMs) have been applied as main learning structure for AVSR systems. HMMs normalize the time axis of the input sequences, and GMMs model the emission probability of each state of the HMM. Two ways of fusing both modalities are used in traditional AVSR: first, early combination (feature fusion) of both feature vectors can be applied~\cite{kratt2004large,neti2000audio}. In some cases algorithms such as Principal Component Analysis (PCA) or Linear Discrimination Analysis (LDA) are used to reduce the dimensionality of those representations. This approach may lead to frame synchronization problems~\cite{kratt2004large,matthews1998features,neti2000audio}. Second, score combination (late fusion) is performed in order to avoid such problems, and even allow for asynchrony in the state sequences in the two streams. In~\cite{luettin2001asynchronous} for example both modalities are analyzed separately and later on the results of both are fused using a bias.

In ~\cite{ngiam2011multimodal,hu2016temporal,mroueh2015deep} present different recent deep-learning approaches to solve the AVSR problem. In~\cite{ngiam2011multimodal} and~\cite{mroueh2015deep}, a joined (audio and video) representations using Deep Neural Networks (DNNs) is learned to perform word and phone recognition respectively. However, no temporal dependence, which is an inherent property of audiovisual speech recognition, is considered. More recently,~\cite{hu2016temporal} present a Recurrent Temporal Multimodal Restricted Boltzmann Machine (RTMRB), which takes into consideration long-term dependencies and outperforms other non-temporal solutions. All approaches explicitly align states with the data and care must be taken in aligning data and setting up experiments.

\subsection{Audio and Video Feature Representation}
\label{sec:featuresvideos}

Several phonetic studies tried to understand which are the most relevant features that can be extracted from the face in order to perform audiovisual speech recognition.  According to~\cite{matthews1998features}, lip position is a considerable source of information when performing visual-only speech recognition. In addition to the position of the lips,~\cite{matthews1998features} state that teeth visibility eases the process of guessing the sound that was produced. Moreover,~\cite{vatikiotis1998moving,vatikiotis1996dynamics,vatikiotis1996characterizing} conduct experiments where it is shown that the entire face provides information about speech.

Traditionally, researchers use different processing and feature extraction methods in order to represent the features explained above. All of them are based on extracting Regions-Of-Interest (ROIs) of each frame where the mouth and other parts of the face (e.g., jaw) are located. Different techniques are used to parametrize the ROIs such as using grey-scale value of each pixel, extracting the variation of the values of each pixel between frames, or parametrize each part of the face using a specific statistical model. As we will discuss in Section~\ref{sec:features}, those techniques have problems since they provide neither a rotation nor a light-invariant feature representation of the area described. 

In the field of deep learning,~\cite{noda2015audio} proposed a MSHMM infrastructure, which uses features extracted from a Convolutial Neural Network (CNN). In addition, as we explain in Section~\ref{secsec:architecture}, ~\cite{ngiam2011multimodal,mroueh2015deep,hu2016temporal} learn a joint feature representation using different DNNs approaches. All those solutions, in turn, are trained on pairs of raw images and the corresponding phoneme labels, which may be unreliable, because of the inherent potential for asynchrony between audio and video (both due to the speech production and due to the technical processes when handling audio-visual content).

\section{Architecture of our system}
\label{sec:pagestyle}

We use the Eesen framework~\cite{miao2015eesen}. The Acoustic Model (AM) is composed of multiple stacked LSTM Networks, and uses CTC as loss-function. This set-up allows to our system to automatically align the sequence of vector representations and the phoneme sequences. It is important to note that the system will output the additional CTC symbol ``blank'' most of the time.

In place of the HMM, a series of three Weighted Finite State Transducers (WFSTs) is used to model the sequence of a symbol and blank states that make up a token (phoneme or a viseme), then the words, and the Language Model (LM) during decoding. 

In our pipeline, four layers of RNNs are connected to build our AM. To provide the ability of learning more complex time sequences we use bidirectional LSTM units~\cite{hochreiter1997long} for our RNN. $K$ possible labels (45 phonemes or 12 visemes in our case) plus a \textit{blank} label that is added in the position 0 compose all the possible output symbols of the network. Let $\mathbf{X}= (\mathbf{x}_{1},...,\mathbf{x}_{T})$ and $\mathbf{z}= (z_{1},..., z_{U})$ with $U\leq T$ be the utterance (audio, video or audiovisual features) and their corresponding label sequence (phonemes), respectively. Thereby, each $\mathbf{x_{i}}$ is a feature vector (audio, visual or audiovisual) and $z_{u} \in \{1,2...K\}$ . The CTC loss function aims to maximize the expression $P(\mathbf{z}|\mathbf{X})$ by optimizing the parameters of the RNN. Since the output of the RNN will be a probability distribution over all possible labels, the last layer of the network is a softmax layer with $K+1$ units (original number of symbols plus \textit{blank}). 

We assume that the probabilities for each time frame are i.i.d. Let $\mathbf{y}_{t}$ be the output probability vector computed for each time frame and $\mathbf{p}= (p_1,..,p_T)$, be a possible output sequence, where $p_t \in \{1,2...K+1\}$. Then the total probability of each possible output sequence of the labels can be computed as:

\begin{equation}
 P(\mathbf{p}|\mathbf{X}) = \prod_{t=1}^{T}y_{t}^{p_{t}}
\end{equation}

We denote all possible $\mathbf{p}$ that can be mapped to a $\mathbf{z}$ as $\Phi{(\mathbf{z})}$. Therefore, the likelihood of $\mathbf{z}$ given an input sequence $\mathbf{X}$ can be described as follows:\

\begin{equation}
 P(\mathbf{z}|\mathbf{X}) = \sum_{\mathbf{p}\in{(\Phi{(\mathbf{z})})}}{P(\mathbf{p}|\mathbf{X})}
\end{equation}

This is the loss function that our RNN aims to maximize.

To allow \textit{blanks} symbols in $\Phi{(\mathbf{z})}$, we add them at the beginning, the end, and between each symbol of $\mathbf{z}$. Consequently, a modified label sequence of length $2U+1$ is to be used to compute $P(\mathbf{p}|\mathbf{X})$. To do so, the well-known \textit{forward-backward algorithm} is used. It computes the probabilities of every past path that ends with a label $u$ at a concrete time $t$ as $\alpha_t-1^{u}$, and the probability of all possible paths that start with label $u$ at time $t$ to the end as $\beta_t^{u}$. Then, the total likelihood of a sequence $\mathbf{z}$ given $\mathbf{X}$ is computed as:
\begin{equation}
 P(\mathbf{p}|\mathbf{X}) = \sum_{u=1}^{2U+1}{\alpha_{t}^{u}\beta_{t}^{u}}
\end{equation}

This is differentiable and can thus be used as objective function.

\section{Data and Features}


\label{secsec:data}

The IBM ViaVoice~\cite{neti2000audio} data set is used to test and train the proposed pipelines. The data set consists of 17111 utterances, which are spoken by 26 different speakers looking directly to the camera with an estimated Signal-to-Noise Ratio of 19.5dB in the audio channel. The data was initially split in 17111 utterances of 261 speakers for training (about 34.9 h) and 1893 of 26 speakers utterance (4.6 hours) for testing. However, we were only able to use 15963 utterances for training and 1840 utterances for testing (see Section~\ref{sec:featuresvideos}). We perform a data augmentation adding white Gaussian noise at 10 different levels of to the original audio signal, which actually have an initial 19.5dB SNR (office noise), creating thus different SNR (40dB to 20dB).

\label{sec:features}


FBank features (40 dimensions), FBank + pitch features (43 dimensions) and Mel Cepstral Coefficients (12 dimensions) are used as audio features in our experiments. Cepstral mean and variance normalization is conducted for robustness, and plus/minus one frame is stacked at the inputs of the neural network.

18 coordinate points that define the inner and outer profile of the mouth shape are extracted using IntraFace ~\cite{de2015intraface}. Afterwards, a position normalization is applied doing an affine transformation to the fixed points (e.g., eye corners) of an average face. Then, a translation to the center of coordinates is performed to the lips contour (inner and outer). Finally, the acceleration and speed of each mouth point is computed. All features described are concatenated to form the visual representation (72 dimensions) of each frame.

A richer representation of the visual modality is achieved describing the mouth landmark points using a scale invariant local description (SIFT)~\cite{lowe2004distinctive}. The original vector of the SIFT descriptors of all the mouth landmarks points has 2304 dimensions. However, the dimensionality of that feature representation is reduced applying a PCA decreasing the number of feature dimensions to 222 (98\% of variance). This information is added to the previous vector to create a more complete representation (294 dimensions) of each frame. IntraFace is not able to process some utterances due to the quality of the data. Therefore, as we already stated in section \ref{secsec:data}, some utterances are removed.

\section{Experiments}
\label{sec:experiments_and_results}
\label{secsec:experiments}

We perform a baseline audio-only recognition experiments with FBank + pitch coefficients and an in-domain language model as used in~\cite{potamianos2001large}, and achieved a WER of 11.8\% in clean conditions using the entire database. In the following experiments a subset of the training and testing set was used, as it is explained in \ref{secsec:data}, some data had to be removed. Also, in order to reduce the language model bias of this setup in the ViaVoice domain, we decided to switch to a more general n-gram language model based on TED talks~\cite{cantab}, which we reduce to the required vocabulary and we use in the following experiments.

\begin{figure}[t]

\includegraphics[width=\linewidth]{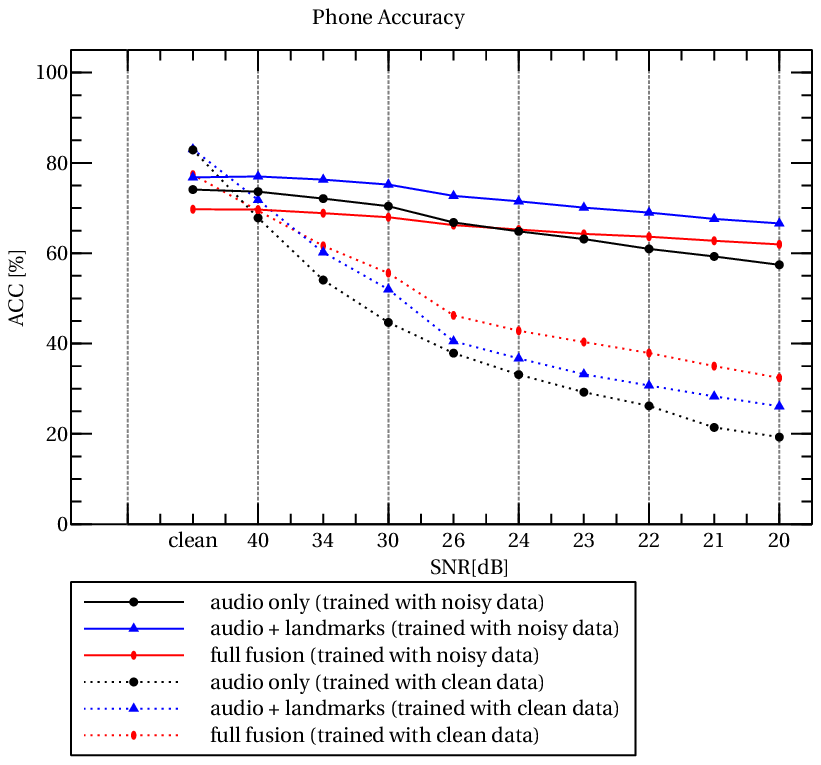}
\caption{Phoneme Accuracies on heldout data, as measured during CTC training.}\label{fig:acc}
\end{figure}

\begin{figure}[t]

\includegraphics[width=\linewidth]{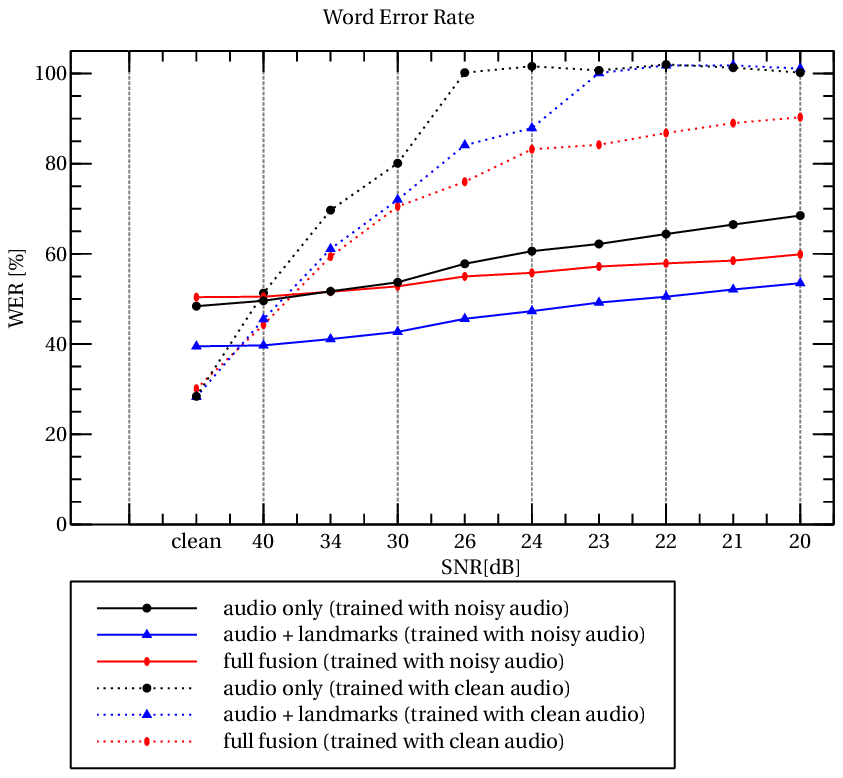}
\caption{Word Error Rates of the different systems. Note that the drop
from 14.9\% WER (Table~\ref{tab:audio}) to 28.7\% WER for the ``clean'' case is due entirely to the general (rather than highly domain-specific) language model.}\label{fig:wer}
\end{figure}

\subsection{Audio Results}

 All features are tested using a 33ms frame rate. This setup is chosen to be as close to the video frame rate as possible for later fusion experiments. As can be seen in Table~\ref{tab:audio}, FBank + pitch leads to better results. This feature is used in the following multi-modal experiment. It is interesting to note that relatively small differences in phone accuracy translate into bigger differences in testing word error rate.

Figure~\ref{fig:acc} shows the phone error rate, Figure~\ref{fig:wer} the word error rate for different noise conditions during testing, of systems trained with clean data only, as well as all noise conditions found in the test data (multi-condition training). ``Clean'' data is very uniform (the recording condition is identical for all speakers), so that multi-condition training does not improve over the baseline in this case. Those results are computed using the reduced dataset explained in section \ref{secsec:data}. However, experiments show that we achieve a 11.7 \% WER using FBank Pitch features with the complete data set.

\begin{table}[t]

\centering

\centering
\begin{tabular}{ l c r }
  \hline
  \textbf{Features} & \textbf{WER} & \textbf{Phone Accuracy} \\ 
    \hline
 \rowcolor[gray]{0.9}  MFCC  &  15.2  \% & 81.5 \%   \\
    \hline
  FBank  &  14.9 \% & 82.7 \%  \\
    \hline
\rowcolor[gray]{0.9} FBank Pitch &  14.4 \% & 83.0 \%  \\
    \hline
\end{tabular}
    
    \caption{Baseline results obtained with different audio features.}\label{tab:audio}
\end{table}

\subsection{Video Results}
\label{secsec:video}

\begin{table}[t]
\centering

\begin{tabular}{l  r }
\hline 
\textbf{Metrics} & \textbf{Viseme Accuracy}\\
\Xhline{2\arrayrulewidth}
  \rowcolor[gray]{0.9}SIFT & 60.8 \% \\ \hline 
 Landmarks   & 40.0\%\\ \hline
 \rowcolor[gray]{0.9} SIFT Landmarks &  63.1 \%  \\ \hline
SIFT  Landmarks Speed Acceleration & 65.7 \%  \\ \hline

\end{tabular}

    \caption{Summary of the results obtained with the different visual feature representations, mapping phonemes to 12 visemes according to~\cite{jeffers1971lipreading}.~\label{tab:viseme_accuracy} }

\end{table}

As can be seen in Table~\ref{tab:viseme_accuracy}, different combinations of visual feature representations have been tested using visemes as target units. SIFT descriptors perform particularly well.

\begin{figure}[t]
\includegraphics[width=\linewidth]{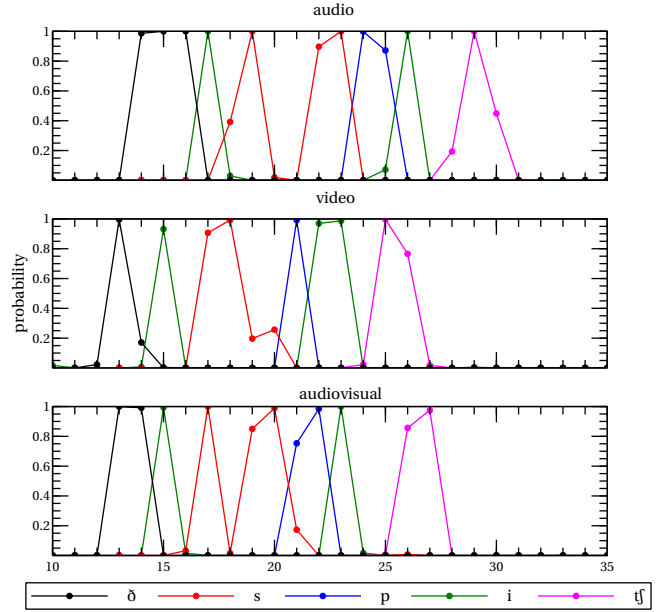}
  \centering
\caption{Averaged position where the RNN of the three systems (audio, video and audiovisual) aligns each phoneme of the words ``the speech''.}\label{fig:alingment}
\end{figure}

\subsection{Audiovisual Results}

Figures ~\ref{fig:acc} and~\ref{fig:wer} show the benefits of training a model using data augmentation techniques, and the benefit of training a multi-modal system on audio and video features.\footnote{Still, because of the homogeneous nature of the data, a model trained specifically for a concrete noise condition will perform better than more general, multi-condition models.} ``Full fusion'' (audio+landmarks+SIFT) models do not perform better than audio+landmarks models at higher SNR, because the high dimensionality of the SIFT feature dominates the audio features, rendering them less useful. We are currently experimenting with further dimensionality reduction techniques to solve that problem.

Figure~\ref{fig:alingment} shows the peaks with which CTC labels the units in each modality, after off-setting technical delays caused by the codec and other factors, for the audio-only, video-only, and audio-visual case. Several conclusions can be extracted from that figure. First, the video signal always precedes audio signal. This finding supports the coarticualation \cite{kent1977coarticulation} and anticipatory coarticulation \cite{bell1982temporal} studies of natural speech production, where is stated that speaker changes the mouth shape before pronounce the following phone. Moreover, from an AVSR point of view, audio-visual models seem to generate better alignments than uni-modal inputs: in most cases, CTC places the position of each phoneme between the audio-only and the video-only position, correcting mis-alignments. Finally, we performed audio-visual training with +/-330ms offset between feature types, without any significant change in WER or PA. This shows that the ``peaky'' structure of CTC is well suited also to multi-modal fusion, and that more detailed studies should be performed in order to investigate the, presumably,\textsl{speaker- and phoneme-dependent} nature of coarticulation \cite{kent1977coarticulation,bell1982temporal}.

\section{Conclusions}
\label{sec:conclusions}

In this paper, we demonstrated that end-to-end Deep Learning can successfully be applied to the problem of audio-visual (multi-modal) speech recognition. Using the CTC loss function and early integration (feature fusion), our system achieves the lowest published word error rate on the large vocabulary IBM ViaVoice database. We show that multi-condition training can be used to improve results on noisy data, and that audio-visual fusion improves results in all conditions, as expected.

More interestingly, the multi-modal setting allows us to reason about the inherent meaning of the ``peaky'' output structure of CTC models, and investigate how their location corresponds to our intuition about the speech production process. 

Reasonable care has been used to tune the models used in these experiments, but further improvements seem possible by exploring more data augmentation strategies, or by further optimizing the CTC training strategy in this relatively low data scenario. Furthermore, neural methods could also be investigated to achieve late fusion.

\vfill\pagebreak

\label{sec:refs}

\bibliographystyle{IEEEbib}
\bibliography{strings,refs}

\end{document}